\documentclass[letterpaper, 10 pt, conference]{ieeeconf}  

\IEEEoverridecommandlockouts                              

\usepackage[acronyms,symbols,index,toc]{glossaries}
\usepackage{graphicx}
\usepackage{url}
\usepackage{hyperref}
\usepackage{listings}
\usepackage{xcolor}
\newacronym{HRI}{HRI}{Human-Robot Interaction}
\newacronym{HHI}{HHI}{Human-Human Interaction}
\newacronym{HCI}{HCI}{Human-Centered Interaction}
\newacronym{LLM}{LLM}{Large Language Model}
\newacronym{LLMs}{LLM}{Large Language Models}
\newacronym{BN}{BN}{Bayesian Network}
\newacronym{BNs}{BNs}{Bayesian Networks}
\newacronym{SCA}{SCA}{Single Condition Assumption}
\newacronym{CPTs}{CPTs}{Conditional Probability Tables}
\newacronym{VVAD}{VVAD}{Visual Voice Activity Detection}
\newacronym{GUI}{GUI}{Graphical User Interface}
\newacronym{CoBaIR}{\emph{CoBaIR}}{\textbf{Co}ntext-\textbf{Ba}sed \textbf{I}ntention \textbf{R}ecognition}
\newacronym{UQ}{UQ}{Uncertainty Quantification}
\newacronym{API}{API}{Application Programming Interface}

\title{\LARGE \bf
CoBaIR: A Python Library for Context-Based Intention Recognition in Human-Robot-Interaction
}

\author{Adrian Lubitz$^{1}$,  Lisa Gutzeit$^{1}$ and Frank Kirchner$^{1,2}$
\thanks{*This work was supported through a grant of the German Federal Ministry of Economic Affairs and Climate Action (BMWi, FKZ 50 RA 2022)}
\thanks{$^{1}$ Robotics Research Group, Department of Computer Science, University of Bremen, 28359 Bremen, Germany
        {\tt\small alubitz@uni-bremen.de, lisa.gutzeit@uni-bremen.de, kirchner@informatik.uni-bremen.de}}%
\thanks{$^{2}$ German Research Center for Artificial Intelligence, 28359 Bremen, Germany
        {\tt\small frank.kirchner@dfki.de}}%
}

\begin{document}

\maketitle
\thispagestyle{empty}
\pagestyle{empty}

\begin{abstract}
        \gls{HRI} becomes more and more important in a world where robots integrate fast
in all aspects of our lives but \gls{HRI} applications depend massively on the
utilized robotic system as well as the deployment environment and cultural differences.
Because of these variable dependencies it is often not feasible to use a data-driven approach
to train a model for human intent recognition.
Expert systems have been proven to close this gap very efficiently.
Furthermore, it is important to support understandability in \gls{HRI} systems to establish trust in the system.
To address the above-mentioned challenges in \gls{HRI} we present an adaptable python library
in which current state-of-the-art Models for context recognition can be integrated.
For Context-Based Intention Recognition a two-layer \gls{BN} is used.
The bayesian approach offers explainability and clarity in the creation of scenarios and
is easily extendable with more modalities.
Additionally, it can be used as an expert system if no data is
available but can as well be fine-tuned when data becomes available.

\end{abstract}

\section{Introduction}\label{sec:introduction}
Our day-to-day lives are becoming increasingly involved with robotic devices.
The industry is currently changing from static robotic environments
to dynamic environments where humans collaborate with robots instead of operating robots.
In private homes, robotic applications like robot vacuums and digital assistants for home
automation are becoming regular household items  \cite{fortunatiIntroductionSpecialIssue2015}.
Although this movement is drastically changing our society,
interactions between humans and robots are very command-driven and unnatural in contrast to \gls{HHI} \cite{yanSurveyPerceptionMethods2014, simExtensiveAssessmentEvaluation2015}.
While \gls{LLMs} like \emph{GPT-3}\cite{brown_language_2020}, \emph{BERT}\cite{devlin_bert_2019}, \emph{LLaMA}\cite{touvron_llama_2023}, and \emph{LaMDA}\cite{thoppilan_lamda_2022} show very promising results in general language understanding,
they have problems with biases, alignment, uncertainty estimation, and most importantly they
lack multimodal understanding of their surroundings which is a key feature for Intention Recognition
needed in modern \gls{HRI} applications \cite{tamkin_understanding_2021}.
In general, data-driven Intention Recognition systems \cite{wangRecurrentConvolutionalNetworks2017,liuIntentionRecognitionPhysical2019} have the
drawback of relying on large and complex (high dimensional) human-robot interaction data.
Because in most scenarios data is not available or impractical to record, data-driven approaches are not suitable for this problem.
Bayesian context-based Intention Recognition is an approach to overcome those limitations
and offer an expert system that can be fine-tuned with data when it becomes available.
Existing research in this direction \cite{pereiraIntentionRecognitionCausal2009,pereiraIntentionRecognitionEvolution2010,panagopoulosBayesianBasedApproachHuman2021,
    kelleyContextBasedBayesianIntent2012,tahboubIntelligentHumanMachineInteraction2006,jainRecursiveBayesianHuman2018} is promising but introduces very complex
network structures which induce the need for the designer of an \gls{HRI} scenario to have a profound knowledge of Bayesian probability theory.
Furthermore, it makes adaptations and re-configuration of the network cumbersome.
To overcome the aforementioned limitations of current systems we propose a two-layer \gls{BN}
for context-based Intention Recognition.
The simple structure enables us to make several optimizations that allow the designer to concentrate on the \gls{HRI} scenario
instead of Bayesian probability theory.

In \cite{lubitzBayesianApproachContextbased2022} we proposed a first concept on how the design for context-based Bayesian Intention Recognition in \gls{HRI} scenarios
can be described in a more compact and intuitive way.
In this paper, we introduce \gls{CoBaIR}, a python software library that comes with the power
to infer intentions from the current context --- context describes every observable aspect in an \gls{HRI} scenario.
Furthermore, \gls{CoBaIR} pays great attention to the design process of \gls{HRI} scenarios.
It provides a configuration format that decreases the number of values that need to be set
during the design process of the Bayesian Network from an exponential to a linear scale.
Additionally, it provides a \gls{GUI} which visualizes the two-layer \gls{BN}
with its weights and offers an intuitive way of configuring it.

This paper starts by pointing out the key challenges in Intention Recognition for \gls{HRI} in Section \ref{sec:challenges}.
In Section \ref{sec:bayesian} we provide a detailed view of the proposed two-layer \gls{BN} structure
and highlight how this structure can make the process of designing \gls{HRI} scenarios easier and faster.
In Section \ref{sec:advantages} we highlight the advantages of the proposed approach.
We point out the most important features of the python implementation in Section \ref{sec:cobair}.
In Section \ref{sec:application} we show an example of how the described python library was used in a research project.
Finally, in Section \ref{sec:conclusion} we conclude with the interpretation of the results and an outlook on future work.

\section{Related Work}\label{sec:related}

During the extensive literature review conducted for this paper,
it was observed that existing intention recognition implementations were often
tightly coupled with specific modalities \cite{jangHumanIntentionRecognition2014,zhangMakingSenseSpatioTemporal2020}
or designed exclusively for particular scenarios \cite{wangVirtualRealityRobotAssisted2020,xingDriverLaneChange2019}.
In some cases, these limitations were found to coexist \cite{zhuWearableSensorsBased2008},
further hindering the applicability of such implementations.
However, as part of our objective to provide a generic framework for
\gls{HRI} within the KiMMI Project,
we recognized the need for a solution that could exhibit high
flexibility and adaptability across various modalities and scenarios.

\subsection{Challenges in Intention Recognition}\label{sec:challenges}
\gls{HRI} depends in many ways on the scenario at hand.
For Intention Recognition we define the following challenges that need to be addressed in order
to implement natural and meaningful \gls{HRI} systems:
\subsubsection*{Hardware constraints}
Just like humans, robots come in all shapes and colors.
More precisely, social robots for \gls{HRI} have different sensing modalities to perceive the human and its surrounding.
While some robots are equipped with stereo-vision or RGBD-camera systems and multiple microphones for echolocation,
as well as LiDAR for navigation, a simple digital assistant may only be equipped with one microphone.
The designer of the \gls{HRI} scenario needs to know which sensory modalities are available for the system in question.
\cite{sheridanHumanRobotInteraction2016, hayesChallengesSharedEnvironmentHumanRobot, jostHumanRobotInteractionEvaluation2020}

\subsubsection*{Application specifics}
Furthermore, the designer needs to know about the application scenario which
can vary from space applications over industrial to domestic applications.
In all of these different application scenarios gestures, voice commands, etc. can mean different things.
\cite{sheridanHumanRobotInteraction2016, jostHumanRobotInteractionEvaluation2020}

\subsubsection*{Cultural differences}
Research on cultural differences in social robotics is an often neglected topic
although social robots will be deployed in multi-cultural place e.g. airports in the future in a more and more globalized world.
One behavior may have different meanings in different cultures.
Therefore the designer must be aware of the cultural differences, and for different cultures,
different \gls{HRI} scenarios must be designed.
\cite{jostHumanRobotInteractionEvaluation2020, nehanivMethodologicalApproachRelating2005}
\subsubsection*{Individual differences}
When we think about developing robots that interact with humans in a very intuitive way
we need to ask ourselves what is intuitive for us.
Intuitive may be slightly different from person to person even within one cultural group.
Humans interact slightly different on an individual level based on their knowledge about the interaction partner.
If the interaction partner is not known a default is chosen which allows for adaptation in the future.
While this behavior is very subtle and unconscious in \gls{HHI}
it is an important factor while designing \gls{HRI} scenarios with the possibility for inter-personal adaptation.
\cite{hayesChallengesSharedEnvironmentHumanRobot, jostHumanRobotInteractionEvaluation2020, nehanivMethodologicalApproachRelating2005}

\subsubsection*{Trust \& Acceptance}
Trust in a robotic system is less scenario specific than the aforementioned challenges and can therefore not as obviously
be integrated into the design process of an \gls{HRI} scenario.
Trust is primarily connected with the human's expectation of the behavior of a robot.
If the robot behaves accordingly to the human's expectations the human can foresee the behavior
and build a model of trust for the robot's abilities.
Secondarily, it is connected with the explainability of a behavior.
If the human is not able to foresee the robot's behavior because it is, e.g. not completely deterministic or too complex to foresee,
the human will seek reasons for the observed behavior.
If reasons can be found trust can still be maintained, while if no reason can be found the trust will decrease immensely.
This model of trust can be very complex in a way that trust exists for specific abilities but not for others.
The trust for specific abilities weighted with their specific importance for the human determines the acceptance in the system.
\cite{hayesChallengesSharedEnvironmentHumanRobot, sheridanHumanRobotInteraction2016, jostHumanRobotInteractionEvaluation2020}
\\
In Section \ref{sec:bayesian} - \ref{sec:application} we illustrate how we address these challenges,
which advantages arise from the proposed approach, how it is implemented,
and how it can be used to model an \gls{HRI} scenario.

\section{Architecture of the two-layer Bayesian Network for Context-based Intention Recognition}\label{sec:bayesian}
\begin{figure}[t]
    \centering
    \includegraphics[width=0.75\linewidth]{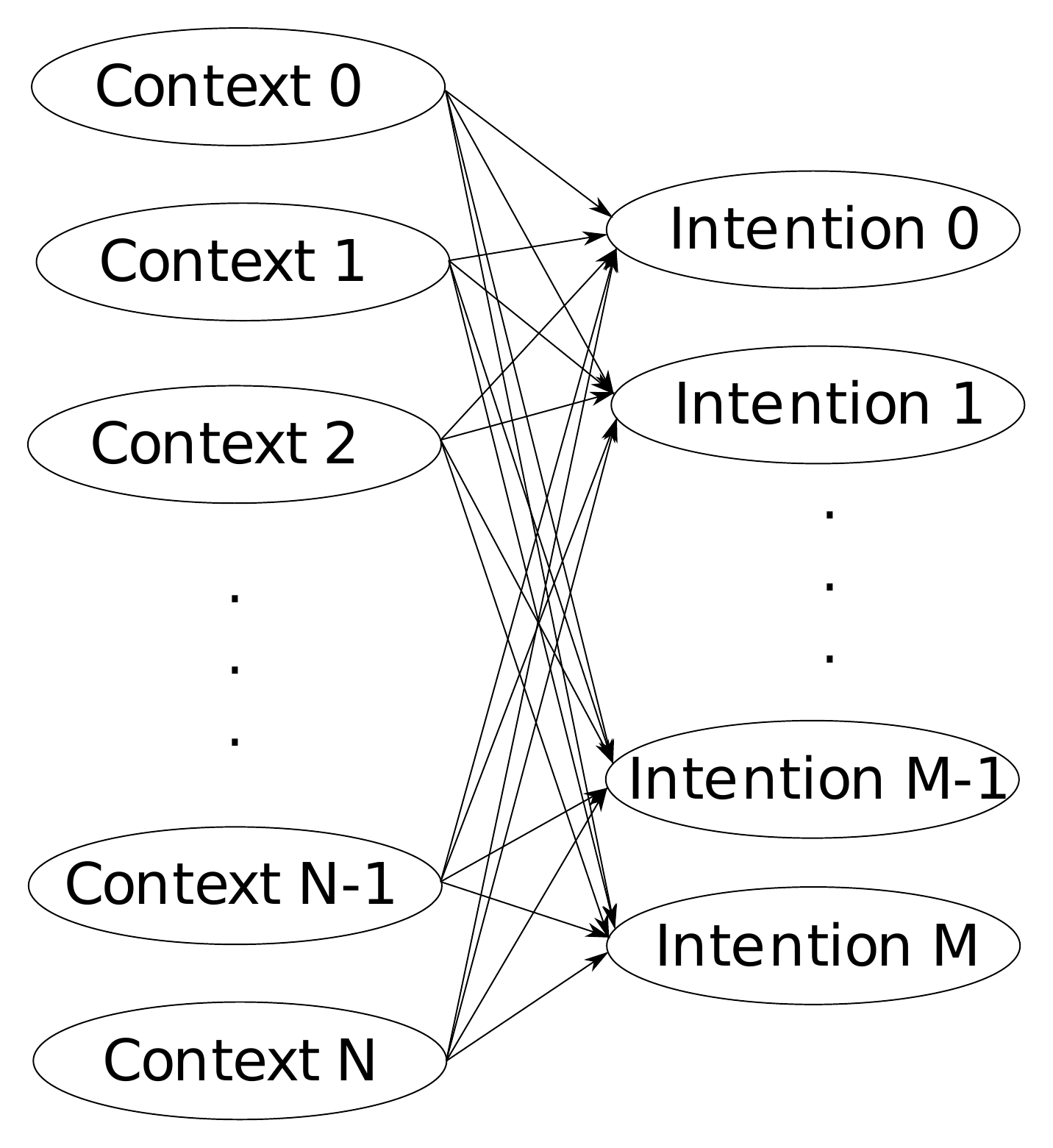}
    \caption{The architecture of the two-layer Bayesian network allows for a high degree of flexibility}
    \label{fig:bayesNet}
\end{figure}

In Section \ref{sec:challenges} we highlighted some of the key challenges in Intention Recognition in \gls{HRI}.
We propose a two-layer \gls{BN} to address these challenges in a computational and data efficient as well as an intuitive way.
The general structure of the \gls{BN} for context-based Intention Recognition is depicted in Figure \ref{fig:bayesNet}.
In \cite{pereiraIntentionRecognitionEvolution2010} and \cite{tahboubIntelligentHumanMachineInteraction2006}
\gls{BNs} with 3 or more layers are proposed to additionally model actions.
We believe the two-layered structure comes with several advantages over using three or more layers.
From a usability point of view, it allows us to assume that the designer of the \gls{HRI} scenario has no
prior knowledge about Bayesian probability in general and \gls{BNs} in specific.

In this way every contexts can be treated in the same way, as an observable phenomenon.
Using further layers would give actions, as suggested by \cite{pereiraIntentionRecognitionEvolution2010}, a special meaning.
\\

This special meaning however is not always valid and
furthermore, it introduces a bias towards actions in the design of \gls{HRI} scenarios.
In cases where external context is of more importance than the performed action, the introduced
bias towards actions can distort the design of the scenario.

In cases where external factors like the time have an influence on the intention
the action itself should not play a predominant role.
The action of grasping a mug could lead to the intention \texttt{make coffee}
but at night the intention becomes more unrealistic and a higher probability
should be given to the intention \texttt{store mug}.
This is a very simple example of how the action bias could lead to the intention \texttt{make coffee}
at night where the correct intention should be \texttt{store mug}.

Modeling relevant observable phenomena as context reduces the action bias and the complexity of the \gls{BN}
and allows the designer of the scenario to concentrate on the specifics of the scenario without the
need for in-depth knowledge about the underlying probabilistic modeling.\\

The final step we took to uncouple probabilistic modeling from the intuitive design of \gls{HRI} scenarios
is to make some assumptions over the given two-layer structure.
These assumptions reduce the probabilistic notion and drastically reduce the number of values
that need to be set by a human expert to describe a scenario.
\subsection*{Independent context and intentions}
The basic assumption is that all intentions are independent of each other as well as all contexts are independent of each other.
This allows for the strict two-layer structure which has only connections between context and intentions as depicted in Figure \ref{fig:bayesNet}.
\subsection*{Binary intentions}
We consider all intentions as binary --- an intention is either present or not.
This allows us to concentrate on the positive (an intention is present) case while designing the scenario and
consider the negative (an intention is not present) case as its complement while calculating the probabilistic model.
This small constraint already cuts the number of values that need to be set by the human expert in half.
Contexts on the other hand can have as many discrete instantiations as needed to create a meaningful \gls{HRI} scenario.
\subsection*{Single Condition Assumption}
We make the \gls{SCA} which implies that every context has an individual
and independent influence on a specific intention.
Using this assumption, we can approximate the conditional probability $P(I_m|C_{1,l}, C_{2, l}, ..., C_{k, l})$
as the average over all single condition probabilities $P(I_m|C_{k,l})$,
where $I_m$ is the \emph{m}-th intention and $C_{k,l}$ is the \emph{l}-th instantiation of the \emph{k}-th context.
This allows us to calculate the conditional probability of the \emph{m}-th intention given the first instantiation for all contexts in the following way:
\begin{equation}
    P(I_m|C_{1,l}, C_{2, l}, ..., C_{k, l}) = \frac{\sum^{k}_{\hat{k}=1} P(I_m|C_{\hat{k}, l})}{k}
\end{equation}
\subsection*{Influence values on a Likert scale}
We generalize the single condition probability  $P(I_m|C_{k,l})$ as an influence value $v_{k, l, m}$ on a six-point Likert scale \cite{likert_technique_1932}
for every context-intention-tuple $(C_{k,l}, I_m)$ to make them more manageable.
The scale is mapped in the following fashion:
$0 \mapsto 0\% ; 1 \mapsto 5\% ; 2 \mapsto 25\% ; 3 \mapsto 50\% ; 4 \mapsto 75\% ; 5 \mapsto 95\% $
\\

The above-mentioned assumptions reduce the amount of values that need to be set by a human expert from an exponential growth given through
\begin{equation}\label{eq:std}
    V(i, j, c, n) = \sum^{j}_{\hat{j} = 1} c_{\hat{j}} + i \times \prod^{i}_{\hat{i} = 1}n_{\hat{i}} \times \prod^{j}_{\hat{j} = 1}c_{\hat{j}}
\end{equation}
to a linear growth given through
\begin{equation}\label{eq:opt}
    V(i, j, c) = (i+1) \times \sum^{j}_{\hat{j} = 1} c_{\hat{j}}
\end{equation}
where $V$ is the number of values to be set,
$i$ is the number of intentions, $j$ is the number of contexts,
$c_j$ is the number of context instantiations for the \emph{j}-th context,
$n_i$ is the number of intention instantiations for the \emph{i}-th intention.
The $\sum^{j} c_j$ in Equations \ref{eq:std} and \ref{eq:opt} describes the amount of \emph{a priori} probabilities for all context instantiations.
The remaining term describes the values needed to fill the \gls{CPTs} manually for Equation \ref{eq:std} and automatically for Equation \ref{eq:opt}.
The \gls{SCA} contributes in a huge way to the ease of designing \gls{HRI} scenarios but ignores cases in which joint probabilities are necessary.
An example could be the handling of voice commands through a robot that is able to estimate directed speech
over \gls{VVAD} as shown in \cite{lubitzVVADLRS3DatasetVisual2023}.
The robot should infer the intention \texttt{pick up tool} with a higher probability if the speech command for picking up a tool was emitted \textbf{AND}
the speech was directed towards the robot than one of them individually.
Only the speech command should have a high probability to infer \texttt{pick up tool} but there is still a chance that the robot was picking up noise
or the speech was not directed towards the robot.
The context of directed speech on the other hand does not have a high probability for any intention individually.
For those special cases, we provide the possibility to set (partially) conditioned influence values
containing multiple contexts that provide more information when combined.
\cite{lubitzBayesianApproachContextbased2022}

While the two-layer \gls{BN} was originally not designed to handle temporal dependencies,
it is possible to model the previously inferred intention as context.
Using a recursive pattern like this, it is possible to model a temporal dependency under the Markov assumption \cite{markov_theory_1953}.

\section{Advantages of a two-layer Bayesian Network for Context-Based Intention Recognition}\label{sec:advantages}
\gls{CoBaIR} uses a two-layer \gls{BN} to represent the dependencies between contexts and intentions.
In this section we want to highlight the key advantages of this structure:

\subsection*{Flexibility}
The biggest advantage of the architecture is its flexibility.
It allows for the usage of any algorithm for context creation, whether it be
probabilistic, heuristic, data-driven or any other approach.
The generated contexts will be used as the input to the two-layer \gls{BN}
which fuses the context information to jointly infer a probability distribution
over all possible intentions.
On the one hand, using \gls{CoBaIR} as an expert system the simplifications explained in Section \ref{sec:bayesian}
allow the human designer to create a scenario in a fast and intuitive manner.
Fast design and adaptation of \gls{HRI} scenarios helps researchers
to concentrate on the specifics of an experiment and therefore reach results faster and more reliably
without any deeper knowledge about Bayesian probability and how to configure \gls{BNs}.
On the other hand the two-layer \gls{BN} can be trained or fine-tuned with data,
which allows to gradually shift from a system trained by an expert to a data-driven approach.
Furthermore, the simple structure of the \gls{BN} allows for the easy removal and addition of contexts.
This makes an iterative prototyping approach, where the \gls{HRI} scenario is build up over time, possible.

\subsection*{Uncertainty Quantification and Explainability}
\gls{UQ} is an often neglected topic, especially in the field vision based tasks \cite{valdenegro-toroFindYourLack2021a}.
The Bayesian approach for Intention Recognition offers the implicit advantage that it comes with a
inherently good \gls{UQ} due to the probabilistic nature of the model.
Additionally, the compact structure of the two-layer \gls{BN} allows users to easily identify
the contexts that played the predominant role in the decision making.
Using this interpretable compact structure we are able to generate explanations to understand the
decisions made by a robot using \gls{CoBaIR} as its Intention Recognition system.
Explainability and \gls{UQ} in a system strongly increases the trust in the system
and therefore the acceptance to use the system in general \cite{sannemanTrustConsiderationsExplainable2020}.

\subsection*{Modularity}
Another advantage of the described architecture is its modularity.
Using a two-layer \gls{BN} to fuse the output of different modules that provide contexts
makes it possible to use existing solutions for context creation, like PAZ \cite{arriagaPerceptionAutonomousSystems2020}
which provides a large variety of models for visual perception, as well as use case specific models that need to be trained from scratch.
Furthermore, it is possible to switch seamlessly between different models on the fly for evaluation and optimization of \gls{HRI} scenarios.

\subsection*{Handling missing input}
While most data-driven approaches have problems handling missing input \cite{ipsenHowDealMissing2020}
and additional data needs to be recorded or artificially generated,
\gls{BNs} provide the advantage of defining \emph{a priori} probabilities for the input.
The \emph{a priori} probabilities for the context can be estimated by an expert or calculated from a few observations.
In this way knowledge about missing input can be incorporated
and during inference time the missing inputs will handled accordingly.
\\

The above-mentioned advantages of the two-layer \gls{BN} highlight why we think a two-layer \gls{BN} is suitable
to enhance the design, inference and explainability of Intention Recognition in \gls{HRI} scenarios.

\section{CoBaIR: a Python Library}\label{sec:cobair}
\gls{CoBaIR} is a python library for \textbf{Co}ntext-\textbf{Ba}sed \textbf{I}ntention \textbf{R}ecognition.
The library allows to create complex \gls{HRI} scenarios in a fast and intuitive manner.
Furthermore, it provides a \gls{GUI} which visualizes the underlying two-layer \gls{BN}
and guides through the configuration procedure.

\gls{CoBaIR} is divided into two parts:
\subsection{Core library}
The core library provides all the key features described in Section \ref{sec:bayesian}
to make the design of \gls{HRI} scenarios intuitive and fast.
It mainly provides the class \texttt{BayesNet} which handles the creation of the two-layer \gls{BN}
from a given configuration.
The configuration is provided in YAML and its fields and format is depicted in Listing \ref{ls:format}.
The format contains the fields \emph{contexts}, \emph{instantiations}, \emph{intentions} and \emph{decision\_threshold}.
\emph{contexts} gives names to the observable phenomena, like \emph{weather}.
\emph{instantiations} are the discrete instantiations of that phenomenon, like \emph{cloudy, rainy, sunny}.
The binary \emph{intentions} are the inferable intentions in the scenario, like \emph{turn on sprinkler}.
All \emph{instantiations} have an a priori probability which needs to be set and furthermore the \emph{instantiations}
have an influence value for each \emph{intention}.
Additionally, there is a field \emph{decision\_threshold} which can be a value between 0 and 1 and
denotes the threshold an intention's likelihood needs to surpass during inference to be considered as the inferred intention.
If the likelihood of the most likely intention is below the threshold, \texttt{None} is returned as the inferred intention.

\begin{lstlisting}[caption=Configuration format for Intention Recognition with \gls{CoBaIR},label=ls:format, basicstyle=\tiny]
    contexts:
      context 1:
        instantiation 1 : float
          .
        instantiation m_1 : float
      context n:
        instantiation 1 : float
          .
        instantiation m_n : float
    intentions: 
      intention 1:
        context 1:
            instantiation 1: int # one out of [5, 4, 3, 2, 1, 0]
            .
            instantiation m_1: int # one out of [5, 4, 3, 2, 1, 0]
        context n:
            instantiation 1: int # one out of [5, 4, 3, 2, 1, 0]
            .
            instantiation m_n: int # one out of [5, 4, 3, 2, 1, 0]
      intention p:
        context 1:
            instantiation 1: int # one out of [5, 4, 3, 2, 1, 0]
            .
            instantiation m_1: int # one out of [5, 4, 3, 2, 1, 0]
        context n:
            instantiation 1: int # one out of [5, 4, 3, 2, 1, 0]
            .
            instantiation m_n: int # one out of [5, 4, 3, 2, 1, 0]
    decision_threshold: float
\end{lstlisting}
The core library provides the means to validate, load and save the configuration.
With the information from the configuration a fully defined two-layer \gls{BN} will be created.
bnlearn \cite{Taskesen_Learning_Bayesian_Networks_2020} is utilized as a backend to handle the two-layer \gls{BNs} and do inference on them.
The \gls{API} is fully documented an publicly available under \href{https://dfki-ric.github.io/CoBaIR/API}{https://dfki-ric.github.io/CoBaIR/API}.

\subsection*{Graphical User Interface}
\begin{figure}[t]
    \centering
    \includegraphics[width=\linewidth]{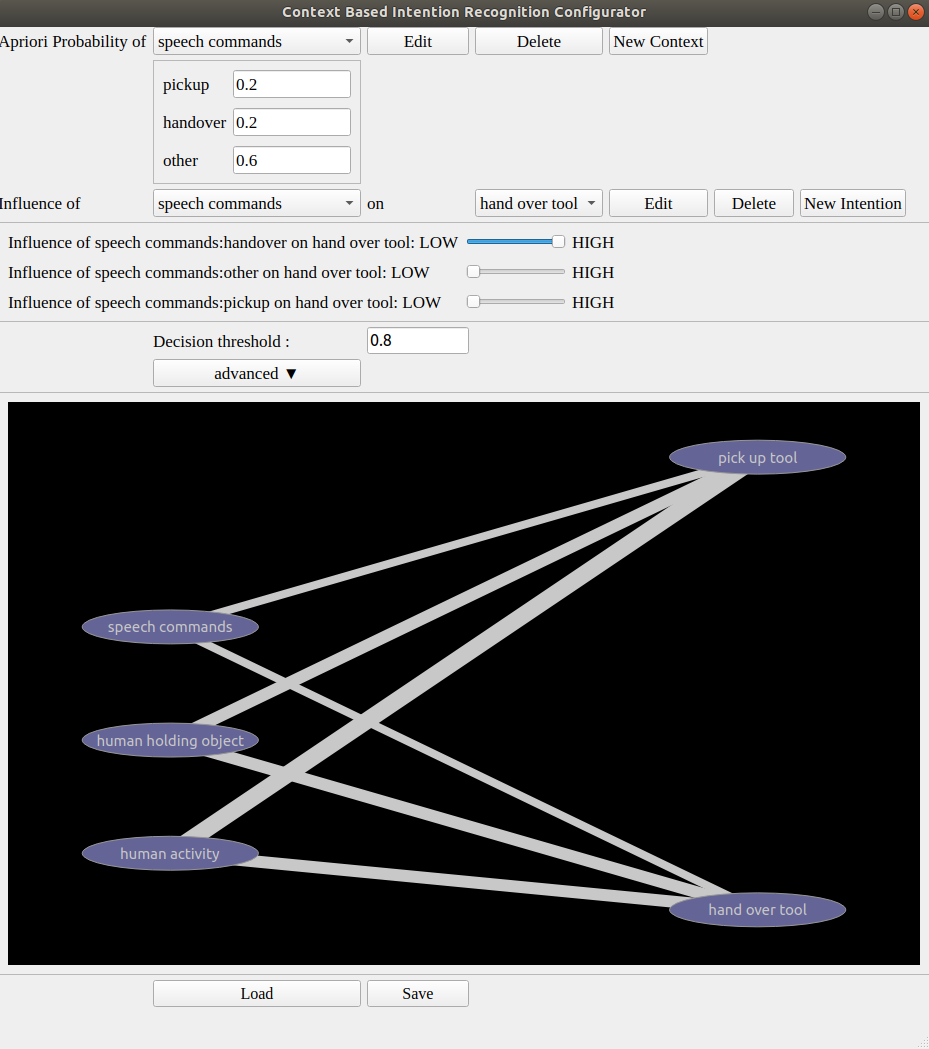}
    \caption{The \gls{GUI} of \gls{CoBaIR} supports the designer of an \gls{HRI} scenario.}
    \label{fig:visualize}
\end{figure}

On top of the optimizations we described in Section \ref{sec:bayesian}, we provide researchers and practitioners
with a helpful \gls{GUI} depicted in Figure \ref{fig:visualize}.
The \gls{GUI} supports the designer to create a configuration
and makes sure it is always valid and provides helpful insights in the case of an invalid configuration.
Furthermore, it visualizes the two-layer \gls{BN} in a live view
which helps to keep a good overview of the configuration.
\\
The complete Python software package is available on \href{https://pypi.org/project/CoBaIR/}{PyPI}.
Additionally, the code is open sourced on \href{https://github.com/dfki-ric/CoBaIR}{GitHub} and open for contributions.

\section{Application of CoBaIR}\label{sec:application}
\begin{figure*}[t]
    \centering
    \includegraphics[height=5cm]{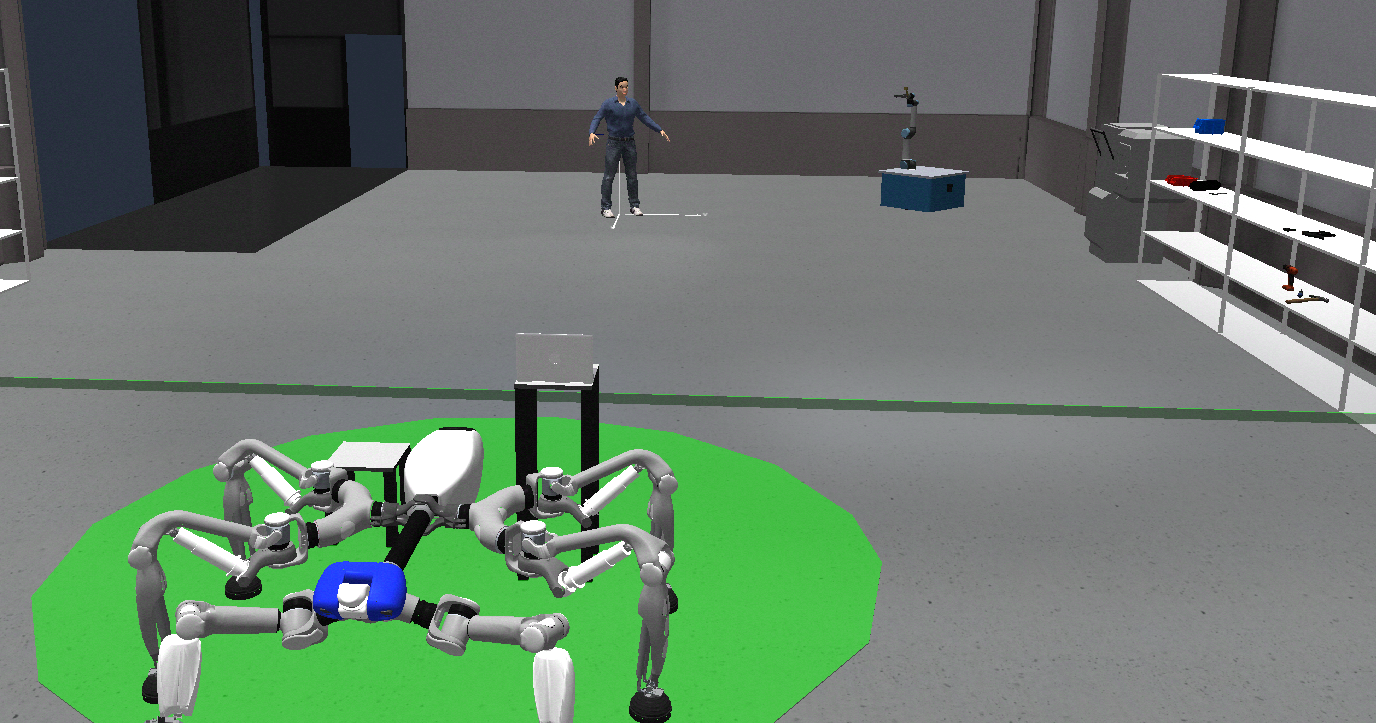}
    \includegraphics[height=5cm]{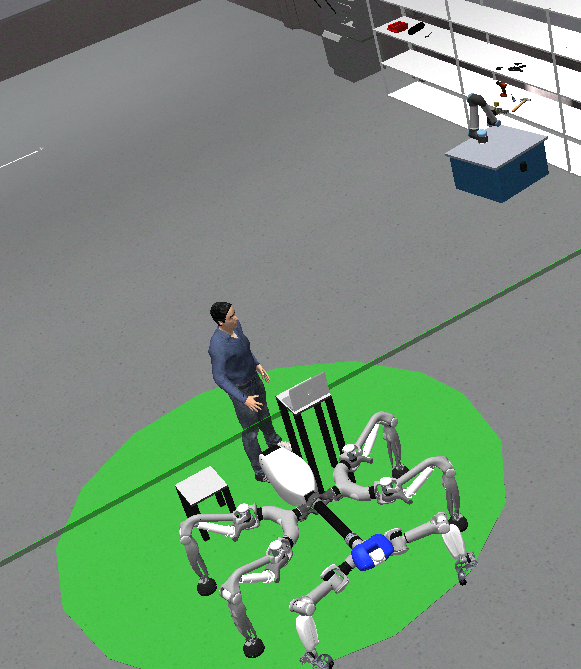}
    \caption{Screenshots of the \gls{HRI} application scenario, in which a simulated human interacts with the UR5 robot mounted on a movable base. The scenario includes a shelf with tools stored in it, a dark area which should be lighted by the UR5 if the human needs support in the dark area, and a work station highlighted in green area where the human can work on the robotic system MANTIS and the UR5 should assist by bringing the desired tool.}
    \label{fig:simulation}
\end{figure*}

We applied \gls{CoBaIR} in an interaction scenario in the project KiMMI SF\footnote{\url{https://robotik.dfki-bremen.de/en/research/projects/kimmi-sf/}}
to infer the intentions of a human operator.
This user controlled a simulated human during the interaction with the
simulated robotic arm from Universal Robot UR5 mounted on a movable base.
The simulated environment is depicted in Figure~\ref{fig:simulation}. The human can be navigated
through the environment, which contains shelves in which different tools are stored, a working
station with an inactive robotic system MANTIS\footnote{\url{https://robotik.dfki-bremen.de/en/research/robot-systems/mantis/}} that should be repaired, and on the left back side a dark area.
Based on the human behavior the robotic system should infer the human's intention in order to react
accordingly. For example, if the human wants to repair the inactive MANTIS robot, the supporting robot should bring the tool which is
stored in the shelf, which is shown on the right side in Figure~\ref{fig:simulation}.

\begin{figure}[t]
    \centering
    \includegraphics[width=0.9\linewidth]{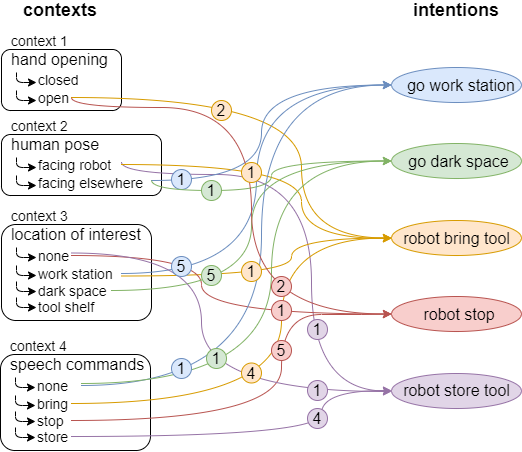}
    \caption{\gls{BN} of the application scenario. Based on four different contexts (left side), five possible human intentions (right side) should be inferred. Arrows indicate the influence of the possible values of the contexts with small numbers representing the assigned influence value. }
    \label{fig:application_net}
\end{figure}

The \gls{BN} used to infer the human intentions is shown in Figure~\ref{fig:application_net}. The human behavior
is measured based on four contexts: \textit{hand opening}, \textit{human pose}, \textit{location of interest}, and \textit{speech commands}, where the first three are determined in the simulation and the speech commands are captured from the operator of the simulation with a
microphone.
Each context is discretized to different values, e.g., \textit{hand opening} can be either \textit{open} or \textit{closed}. This is shown
on the left side of Figure~\ref{fig:application_net} for each context. In the presented scenario, based on the given contexts, the following five intentions should be inferred:
1. \textit{go work station}, i.e., the human wants to go to the work station;
2. \textit{go dark space}, i.e., the human want to go to the dark area;
3. \textit{robot bring tool}, i.e., the human wants the robot to bring the tool stored in the shelf;
4. \textit{robot stop}, i.e., the human wants the robot to stop its current action;
5. \textit{robot store tool}, the human wants the robot to store the tool back to the shelf.

Using \gls{CoBaIR} and the included \gls{GUI} shown in Figure~\ref{fig:visualize}, the \gls{BN} shown in Figure~\ref{fig:application_net} could easily be designed and optimized for the described scenario. With the resulting \gls{BN} all human intentions could be reliably inferred and the correct reactions of the UR5
could be triggered to realize a successful interaction between the human and the robotic system.

During the KiMMI SF project we followed an agile development process
which led to multiple incremental as well as complete changes in the design of the \gls{HRI} scenario.
\gls{CoBaIR} enabled us to incorporate these changes fast and effectively in the development process.

\section{Conclusion and future work}\label{sec:conclusion}
We presented the python library \gls{CoBaIR} in this paper.
We demonstrated that the concept from \cite{lubitzBayesianApproachContextbased2022} using a two-layer \gls{BN}
with the assumptions highlighted in \ref{sec:bayesian} can be effectively implemented.
Additionally, we provided a \gls{GUI} to visualize and guide the design process for \gls{HRI} scenarios.
In Section \ref{sec:application} we showed that \gls{CoBaIR}
was successfully used in the KiMMI SF project.
We advise practitioners to use the state-of-the-art models for context recognition
from the open source library PAZ \cite{arriagaPerceptionAutonomousSystems2020}
which is constantly updated with perception models for autonomous systems.
In the future we plan on providing tutorials and examples on how to use \gls{CoBaIR} with PAZ.
While it is theoretically possible to train or fine-tune the two-layer \gls{BN}
with data from the scenario,so far the fine-tuning and data-driven training from scratch was not tested.
In future works we want to investigate the capability of data-driven training and fine-tuning within \gls{CoBaIR}.
Additionally, we want to quantify the effect in terms of quality and speed
which \gls{CoBaIR} has on the design of \gls{HRI} scenarios, conducting user studies
comparing the time and cognitive load to create a \gls{HRI} scenario using our solution
in contrast to creating a \gls{HRI} scenario using \gls{CPTs} for the creation of the \gls{BN}..
We will be incorporating \gls{CoBaIR} in future projects and by making \gls{CoBaIR} open source
we hope to provide a helpful tool for researchers and practitioners in \gls{HRI} all over the world.
Everyone is welcome to use, give feedback and contribute to \gls{CoBaIR} through \href{https://github.com/dfki-ric/CoBaIR}{GitHub}.







\section*{ACKNOWLEDGMENT}
This work was supported through a grant of the German Federal Ministry of Economic Affairs and Climate Action (BMWi, FKZ 50 RA 2022).


\bibliographystyle{IEEEtran}
\bibliography{references}

\end{document}